\documentclass[conference]{IEEEtran}
\IEEEoverridecommandlockouts
\usepackage{cite}
\usepackage{amsmath,amssymb,amsfonts}

\usepackage{relsize}
\usepackage{algorithmic}
\usepackage{graphicx}
\usepackage{textcomp}
\usepackage{xcolor}
\usepackage{graphicx}
\usepackage{multirow}
\usepackage[font=small]{caption}
 
\def\BibTeX{{\rm B\kern-.05em{\sc i\kern-.025em b}\kern-.08em
    T\kern-.1667em\lower.7ex\hbox{E}\kern-.125emX}}
\begin{document}

\title{Edge Attention Module for Object Classification\\
{\footnotesize \textsuperscript{}}
\thanks{\textit{Under review in IJCNN'25}}
}
\author{\IEEEauthorblockN{Santanu Roy}
\IEEEauthorblockA{\textit{Dept. of Computer Science} \\
\textit{NIIT University}\\
Jaipur, India \\
santanuroy35@gmail.com}
\and
\IEEEauthorblockN{Ashvath Suresh}
\IEEEauthorblockA{\textit{Dept. of Computer Science and Engg} \\
\textit{Christ (Deemed to be University)}\\
Bangalore, India \\
ashvath.suresh@btech.christuniversity.in
}
\and
\IEEEauthorblockN{Archit Gupta}
\IEEEauthorblockA{\textit{Dept. of Computer Science} \\
\textit{NIIT University}\\
Jaipur, India \\
gupta.archit2001@gmail.com
}
}
\maketitle

\begin{abstract}
A novel ``edge attention-based Convolutional Neural Network (CNN)'' is proposed in this research for object classification task. With the advent of advanced computing technology, CNN models have achieved to remarkable success, particularly in computer vision applications. Nevertheless, the efficacy of the conventional CNN is often hindered due to class imbalance and inter-class similarity problems, which are particularly prominent in the computer vision field. In this research, we introduce for the first time an ``Edge Attention Module (EAM)'' consisting of a Max-Min pooling layer, followed by convolutional layers. This Max-Min pooling is entirely a novel pooling technique, specifically designed to capture only the edge information that is crucial for any object classification task. Therefore, by integrating this novel pooling technique into the attention module, the CNN network inherently prioritizes on essential edge features, thereby boosting the accuracy and F1-score of the model significantly. We have implemented our proposed EAM or 2EAMs on several standard pre-trained CNN models for Caltech-101, Caltech-256, CIFAR-100 and Tiny ImageNet-200 datasets. The extensive experiments reveal that our proposed framework (that is, EAM with CNN and 2EAMs with CNN), outperforms all pre-trained CNN models as well as recent trend models ``Pooling-based Vision Transformer (PiT)'', ``Convolutional Block Attention Module (CBAM)'', and ConvNext, by substantial margins. We have achieved the accuracy of 95.5\% and 86\% by the proposed framework on Caltech-101 and Caltech-256 datasets, respectively. So far, this is the best results on these datasets, to the best of our knowledge. All the codes along with graphs, and their classification reports are shared on an anonymous GitHub link: 
{https://anonymous.4open.science/r/Object-Classification-7BE5}
\end{abstract}


\section{Introduction}
{I}n recent years, significant progress has been made in the field of object classification tasks, driven by advancements in Convolutional Neural Networks (CNNs) and emerging technologies such as Vision Transformer (ViT) [1]. The concept of ViT has come from the notion of self-supervised learning, which is inspired by the field of Natural Language Processing (NLP) [2]. 
Despite the triumph of ViT in the computer vision field over traditional CNN models, ViT has not yet completely supplanted CNN architectures. Some researchers [3,4] have discovered that CNN models can still surpass ViT if it is trained in an intelligent way. This is because CNN model architectures have multi-scale hierarchical structure [5], where the Max-pooling layer is deployed interchangeably after each convolutional block. These hierarchical structures enable the model to capture multi-scale features [5], thus, extracting a diverse range of patterns that are essential for object classification task. Swin Transformers [6] have integrated both of the principles of CNN and ViT, in order to have a hierarchical structure in the Vision Transformer. Recently Byeongho Heo et al. [7] have proposed a Pooling-based ViT (PiT), which incorporates pooling layers in the ViT model. This achieves two main objectives: (I) Their model architecture now has fewer tokens (or patches), thereby significantly reducing the number of trainable parameters and potentially addressing overfitting issues for small or imbalanced datasets, (II) It incorporates multi-scale hierarchical structure into their ViT model, which is crucial for computer vision tasks. Many more extensive research can be found in [8]-[14] about numerous architectures of ViT.

J. Hu et al. [15] introduced the Squeeze Excitation Network (SE-Net) which is a notable breakthrough in the field of computer vision, known for providing channel attention on top of CNN model. This SE-Net block starts with a concatenation of Global Average Pooling (GAP) [16] and Global Max pooling, thus, diminishing the spatial dimension to 1$\times$1 and solely focus on channels. Therefore, by incorporating this SE-Net block into a CNN model, it gives more attention towards channels, thus, boosting the overall performance of the model. S. Woo et al. [17] incorporated the Convolutional Block Attention Module (CBAM) into a CNN model, which integrates both channel attention and spatial attention. Their channel attention module is inspired by SE-Net, while their spatial attention module begins with the concatenation of Max-pooling and Average-pooling, followed by a single 7$\times$7 filter. Therefore, it automatically provides more focus in spatial domain on larger scale. As a consequence, it significantly enhances the generalization capability of the CNN model. Overall, we have found that, most of the researchers [18-20] have deployed attention mechanisms in the spectral dimension, rather than spatial dimension. In our knowledge, untill now, no effective ``Attention Mechanism'' has been invented that operates in the spatial domain. The reason behind this is that no powerful pooling technique is available for providing attention only towards imperative features like edges, in the spatial domain. In this research, we have invented a very powerful pooling technique called ``Max-Min pooling'', which can directly extract prominent edges from the images. These are salient features for the object classification task [21]. Thereby incorporating this novel pooling technique into an attention module, it automatically focus on the valuable ``edge'' features. Consequently, it acts as an ``Edge Attention Module (EAM)'' that accelerates the loss convergence and inherently addresses the class imbalance issues and other challenges for object classification.

Still now we have encountered only a limited number of research papers that directly focus only on object classification tasks, particularly in the perspective of class imbalance. B. Xiao et al. [22] have introduced an innovative pooling block to enhance the efficacy of lightweight CNN models such as MobileNet-V2 and SqueezeNet.  
Their pooling block encompasses one 2$\times$2 Max-pooling layer, followed by a 1$\times$1 convolution layer and a Batch-normalization layer. The authors asserted that integrating this pooling block into lightweight CNN models, can significantly accelerate the convergence rate, thereby automatically alleviating the class imbalance problem to some extent. J. Pan et al. [23] have proposed a channel-spatial hybrid attention module in order to resolve challenges in Caltech datasets. Subsequently, they have incorporated a novel nearest neighbor interpolation block in order to resize or up-sample the images. Y. Li et al. [24] have proposed a novel adaptive batch-normalization in order to enhance the generalization ability of the Deep CNN models. Many more research on object classification can be found from [25] to [29].

For our experiments, we utilized total four widely recognized datasets, that are Caltech 101 [30], Caltech 256 [31], CIFAR-100 [32] and Tiny ImageNet-200 [33]. Caltech 101 dataset comprises of 10,000 images across a diverse range of classes (total 102), while Caltech 256 dataset extends the challenge with a more extensive set of 60,000 images across 257 classes. Similarly, CIFAR-100 dataset is also the extended version of CIFAR-10 dataset where the number of classes is 101. Moreover, we have prepared one more dataset, that is, Tiny ImageNet-200. In the conventional Tiny-ImageNet dataset, there were 200 number of classes, each containing 500 images. However, due to limited GPU resources (as the implementation is being carried out on Kaggle), we were compelled to reduce the number of images per class. Therefore, we performed under-sampling on this standard dataset to create a relatively smaller dataset, having only 100 images per class. We call this dataset ``ImageNet-200''. Each of these datasets has unique challenge, both of the Caltech datasets are severely class imbalanced. In the Tiny ImageNet dataset there is higher intra-class variance [34] as well as low inter-class variance, which makes the object classification task challenging. On the other hand, CIFAR-100 dataset has been forcefully extended from CIFAR-10, thus, it has higher inter-class similarity [34], that means, many of the images from different classes have similar kinds of statistics. Therefore, it is likely that there will be huge number of miss-classification. This is to clarify that we have chosen both Caltech-101 and Caltech-256 datasets because until now no researchers have resolved the class imbalance problem from these datasets in a generalized way, to the best of our knowledge. Furthermore, this is to notify that MS-COCO [35], PASCAL VoC [35], and iNaturalist datasets [36] are also very popular and having  class imbalance problems, however, their objectives differ as they focus on object detection where multiple objects may be present in a single image. Hence, these datasets have not been considered in this research. 

\vspace{0.1cm}
The contributions of this paper are presented in the following: 
\begin{enumerate}
\item In this research, a novel ``Edge Attention Module (EAM)'' is proposed in which Max-Min pooling is incorporated in the pre-trained CNN model. This module can provide attention towards the most salient features for object classification that are edges, thereby further boosting the accuracy of the model. 

\item We have shown that this EAM block can be leveraged on the CNN model in a very flexible way. For instance, this EAM can be incorporated twice (or, more than that) via parallel branches, as per user choice. 
\item  In order to prove the validity of the proposed framework, we have tested our model on four versatile recognized datasets. Moreover, we have performed a 5-fold cross validation experiment on Caltech-101 dataset for ``DenseNet-121$+$EAM'' framework.
\item Subsequently, we have added GradCam heat map diagrams which further validate our theory on ``Edge Attention Module''.
\end{enumerate}

\begin{figure*}[h]
		\centering
		\includegraphics[width=17.3cm,height=9.7cm]{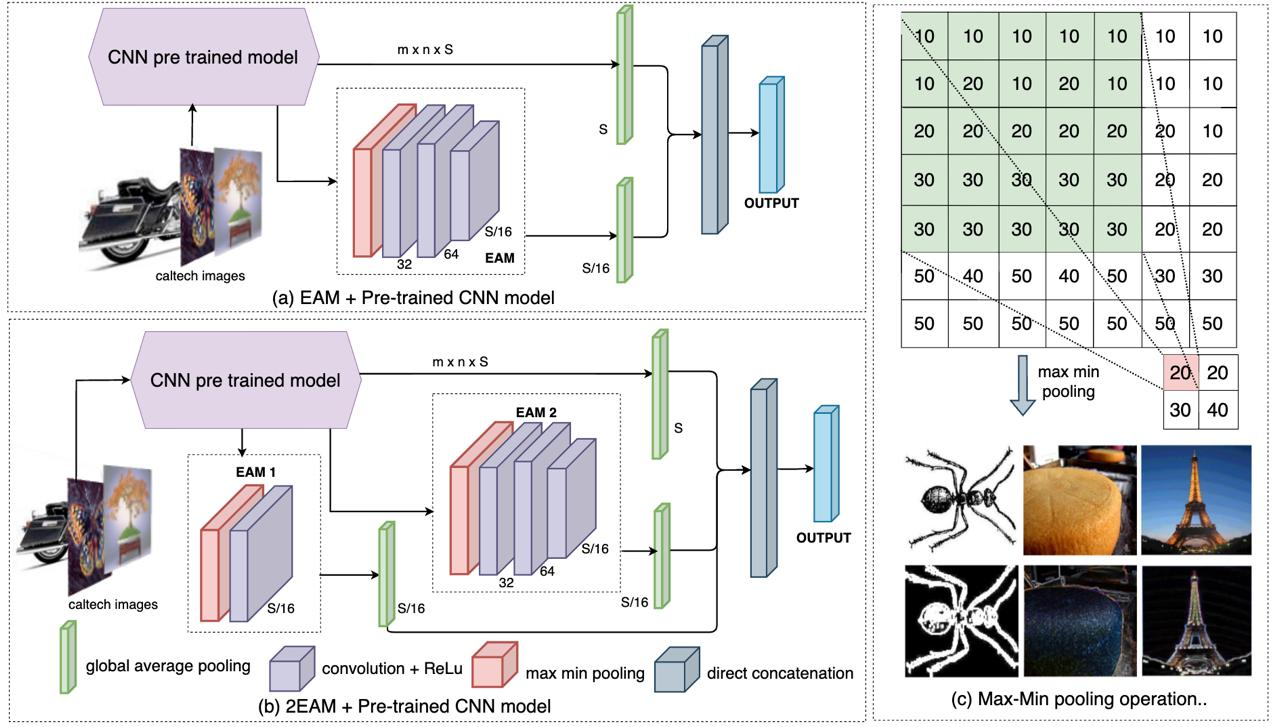}
		\caption{Proposed Frameworks with Edge Attention Module (EAM): a. EAM + pre-trained CNN, b. 2EAM + pre-trained CNN, c. Operation of Max-Min pooling, in the lower image the first row represents the original image and second row represents the processed image after passing the original images through Max-Min pooling layer}
	\end{figure*}
\section{Methodology}
The entire methodology of the proposed framework can be further divided into three parts: (A) Edge Attention Module and Its mathematical analysis, (B) Properties of the proposed Edge Attention Module, (C) Training Specifications.

\subsection{Edge Attention Module and Its mathematical analysis}
An edge attention module (EAM) is incorporated into standard pre-trained CNN model via a parallel branch. This EAM appears like a bridge, originating from the output of $2^{nd}$ last convolutional block of the pre-trained CNN model and it concatenates with the main base model through incorporation of Global Average Pooling (GAP) layer [16], as illustrated in Fig.1a. EAM block consists of a Max-Min pooling layer, followed by $3$ consecutive convolutional layers, as depicted in Fig.1a. This Max-Min pooling is entirely a novel pooling technique, first time proposed in this research. From Fig.1c it is observed that this pooling technique is not like other traditional pooling techniques such as Max-Pooling or Average Pooling. Fig.1c shows that it utilizes a 5$\times$5 window (of stride=2) and computes the total deviation of the intensity values (i.e., maximum minus minimum value) from it. Hence, this is analogous to taking the first difference of intensity values [37] inside this window, which is directly correlated to the first-order derivative or edge information inside the image. Therefore, this Max-Min pooling layer has the ability to extract only the edge information present inside the image, while down-sampling the image by half (/2). These edges are the most essential features for any object classification [38] or recognition task. As a consequence, this EAM block provides additional edge information to the neural network and due to that it further pushes the efficacy of the neural network substantially. \textbf{A detailed analysis of Max-Min pooling with example, is presented in the appendix-I.} 

Generally in digital images, inside a very small (2$\times$2) window, it is likely that the neighbor pixels have very similar intensity values. In other words, they do not deviate much from a particular intensity value $c_i$, even though there are edges present in the image.
Now let's analyze the Max-Min pooling method with 5$\times$5 pool size and stride 2. Let's consider, $c_i$ is the most frequent pixel intensity value inside $i^{th}$ 5$\times$5 window. Let's assume, all the pixel values inside that 5$\times$5 window vary from $c_i-v_{i,min}$ to $c_i+v_{i,max}$. This $v_{i,max}$ and ${v_{i,min}}$ are the maximum deviation of intensity value inside the 5$\times$5 window with respect to $c_i$, in positive and negative direction respectively. Here, $c_i$ and $v_{i}$ are not constant values, their values may differ in each window based on the image statistics.

Now, any Max-pooling operation [5] with 2$\times$2 pool size, can be thought of as a function $F_{m}(\hspace{0.1cm})$ that performs down-sampling (/2) operation on the main input tensor (image) and can be represented by the following equation (1). 
\begin{equation}
		{(F_{m}{(I(x,y))}_{S\times 
 S})}_{2\times 2|2}={I{(max(x,y)}_{2\times 2})}_{\frac{S}{2}\times \frac{S}{2}}
\end{equation}

Here, $I(x,y)$ is the intensity value at spatial domain $(x,y)$. The input tensor size after applying Max-Pooling layer $F_{m}(\hspace{0.1cm})$ is reduced from S$\times$S to $\frac{S}{2}\times\frac{S}{2}$, because stride is 2 here. 

On the other hand, the proposed Max-Min pooling can also be thought of as a function $F_{mn}(\hspace{0.1cm})$ shown in Equation (2). 
\begin{equation*}
		{(F_{mn}{(I(x,y))}_{S\times S})}_{5\times 5|2}=
		{I({max(x,y)}_{5\times 5}}-
  \end{equation*}
\begin{equation}
  {{min(x,y)}_{5\times 5})_{\frac{S}{2}\times \frac{S}{2}}}
\end{equation}
The meaning of this pooling operation is that, first the maximum and minimum values are computed in each 5$\times$5 window of input image, thereafter, the minimum value is subtracted from the maximum value. The other components in the equation (2) are exactly same as equation (1), except the pool size is changed from 2$\times$2 to 5$\times$5 . 

In a conventional Max-pooling operation $F_{m}(\hspace{0.1 cm} )$ with 5$\times$5 window size and stride 2, can be presented by equation (3). let's assume here, total $M$ number of windows are utilized. 
\begin{equation}
		Hence, \hspace{0.3 cm} {(F_{m}{(I(x,y))}_{S\times S})}_{5\times 5|2}={\sum_{i=1}^{M}(c_{i}+v_{i,max})}
\end{equation}

Because maximum intensity value inside $i^{th}$ patch (in 5$\times$5 window) is $c_i+v_{i,max}$. Similarly, a Max-Min pooling operation $F_{mn}(\hspace{0.1 cm} )$, can be represented by equation (4). It is likely that inside $i^{th}$ patch of Max-Min pooling, it will subtract the minimum value from its maximum value. Thus, $c_i$ value will get (approximately) nullified and eventually, this can be represented by the equation (5). 
\begin{equation}
		 {(F_{mn}{(I(x,y))}_{S\times S})}_{5\times 5|2}={\sum_{i=1}^{M}(c_{i}+v_{i,max}-c_{i}+v_{i,min})}
\end{equation}
 \begin{equation}
or, \hspace{0.2 cm} {(F_{mn}{(I(x,y))}_{S\times S})}_{5\times 5|2}\approx{\sum_{i=1}^{M}(v_{i,max}+v_{i,min})}
\end{equation}

Comparing two equations (3) and (5), it can be concluded that Max-Min pooling only can track the edge information (i.e., maximum deviation in positive direction$+$maximum deviation in negative direction with respect to $c_i$), rather than extracting maximum (or, average) intensity information. Therefore, its working function is entirely different from the traditional Max-Pooling [5] or Average-Pooling operation. Furthermore, from Fig.2 it can be verified that a Max-Min pooling operation with 5$\times$5 pool size, can extract only the edge components of the image very efficiently. 

\subsection{Properties of the Edge Attention Module}
Some additional important properties of the proposed EAM framework are given in the following:
\begin{enumerate}
        \item The reason behind choosing the pool size of the proposed Max-Min pooling 5$\times$5 is that, in a smaller window (2$\times$2), the pixel intensity values of neighboring pixels in a digital image are likely to be very similar. Hence, there is a risk that maximum values and minimum values might eliminate each other out in Max-Min pooling and eventually, after down-sampling(/2) all the pixel intensity values might be reduced to zero (or, very closed to zero value). Therefore, a slightly larger window (5$\times$5) is utilized in the proposed Max-Min pooling method in order to capture at least some intensity variation (in terms of edges). Furthermore, this is evident from Fig.2 that, Max-Min pooling with pool size 5$\times$5 can have more edge strength than that of pool size 2$\times$2. 
        \item However, due to focusing only on the prominent edges, it may wash out some important features. For instance, as observed in Fig.2 (second row), if the object is associated with a homogeneous region (like the cake shown figure), then Max-Min pooling may lose some important information about the foreground texture.
        \item Another limitation of this Max-Min pooling is that it can be slightly sensitive to noise. Because there may be intensity variation due to noise, rather than edges in the digital images. Then this noise might be also captured along with this edge information in the output of the Max-Min pooled images.
        \begin{figure}[h]
		\centering       \includegraphics[width=9cm,height=8.3cm]{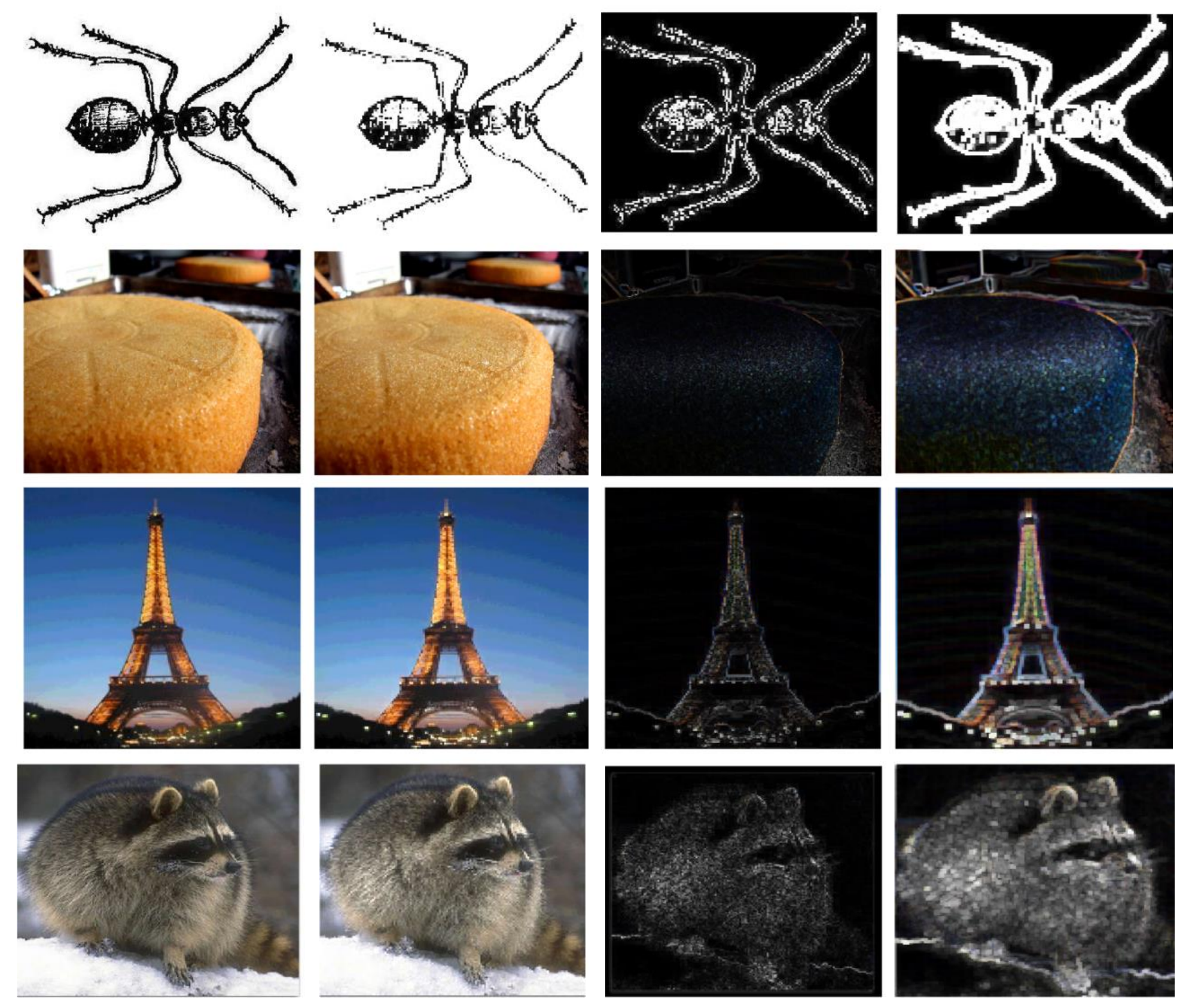}
		\caption{First column represents original images from Caltech, 2nd column represents Max-pooled images (with pool size 2 and stride 2), 3rd column represents Max-Min pooled images (with pool size 2 and stride 2), 4th column represents Max-Min pooled images (with pool size 5 and stride 2), code is given in supplementary material}
	\end{figure}
        \item To overcome the aforementioned limitations (point 2-3), we have deployed this new pooling technique only through a parallel concatenation (not in the main-base model in series), so that it can only act as a catalyst that provides some additional information about the edge of the objects. As a consequence, this EAM block enables the CNN model to pay more attention to the salient features (edges), and reduces the solution space for the object classification task. This leads to faster convergence and improves the accuracy of the model.
        \item For all pre-trained CNN frameworks, we have chosen the same ratio between the final spectral dimension of the base model and that of the attention module. Specifically, this ratio has been chosen empirically as 16:1. This implies that, the final decision of the classifier is dependent mostly on the extracted features coming from the base model. Only a minimal portion (1:16=0.0625) of distinctive (edge) features coming from the attention block is considered for the final decision of the classifier. Hence, the overall decision of the classifier won't depend too much on this Max-Min pooling layer. This approach ensures that the proposed framework is not overly sensitive to noise or information loss, thereby alleviating the limitations of the Max-Min pooling layer. 
\end{enumerate}

In this research, we have proposed Edge Attention Module (EAM) in a very flexible way. For instance, Fig.1b presents a pre-trained CNN model integrated with 2EAMs, that means, EAM block has been utilized twice here. The $1^{st}$ EAM is employed from the output of $3^{rd}$ last convolutional block to direct GAP, and the EAM-2 is employed from output of $2^{nd}$ last convolutional block to GAP, as illustrated in Fig.1. Eventually these GAPs are concatenated with the output of base model. Design perspective we have observed that there is a trade-off. If the challenge is more in a particular dataset, one can incorporate higher number of EAMs, in order to further push the accuracy of the model. However, increasing the number of EAMs may increase the complexity in the model a bit (because Max-Min pooling pool size is 5$\times$5), which may lead to substantial over-fitting in the model performance. Therefore, one should be very careful in choosing the number of EAM blocks in the CNN model. If the model complexity is already too high, avoiding some of the EAMs from the framework can be a good design of choice and vice-versa. The user has to come up with appropriate number of EAMs in the framework according to their own choice, thereby, making this design very much flexible and user-friendly.

This is to clarify that, in this EAM framework, the only essential component is Max-Min pooling layer (with stride=2) that extracts only edges. The other components like 3$\times$3 convolutional layers are optional. Only the final convolutional layer is important, as the edge attention weightage can be regulated by adjusting the number of filters in this layer. Therefore, these convolutional layers (in EAM block) can be altered or skipped based on the problem complexity and (user) requirements. That's why, it is noticed in Fig.1b that we have not deployed any other convolutional layer rather than the last layer, in the EAM-1. Because the model complexity was already higher according to our analysis. In this research, two frameworks (CNN+ EAM and CNN+ 2EAMs) have been proposed for the object classification task. This is to clarify that it is not always the case that 2EAMs will outperform the EAM framework. It depends on the model complexity, challenges in the dataset and many more factors. In the results and analysis part, we have shown that, after incorporating EAM or 2EAM branch, there is consistent improvement for any CNN model throughout all the four datasets. 

\subsection{Training Specifications and Implementation of models}\label{AA}
All the CNN models have been built using Keras sequential API, only ViT model is performed on the platform of Pytorch. Some of the training specifications are given below: 
\begin{itemize}
    \item Caltech-101 and Caltech-256 both the datasets are divided into training, testing and validation sets with a ratio of 65\%-15\%-20\%. This splitting is done in a stratified way [39], which is feasible for class imbalance problem. However, for ImageNet-200 and CIFAR-100 datasets were not imbalanced, therefore, we have done the conventional splitting method that is 80\% training and 20\% testing. 
    \item All the images are resized into 224 x 224, prior to feeding them into CNN models. 
    \item A batch size of 6 is utilized while training all the CNN models, except ConvNext [40]. We had to set so much less batch size, due to deploying two EAMs the model's complexity was increased a bit. In the contrary, due to deploying the latest advancements, ConvNext model converges much smoother than any other model. We have empirically found that batch size of 32 is optimal for ConvNext. 
    \item In all pre-trained CNN models, no fully connected layer (or, dense layer) is taken into account, in order to avoid over-fitting problem.
    
    \item We have implemented the proposed framework with the Early Stopping (ES) criteria of 3,5, and 10 epochs patience, monitored on validation loss. Out of all these experiments, only the best outcomes from these experiments are reported in the Tables. Here, we explored a variety of ES criteria in order to investigate the actual effectiveness of the proposed attention module. For most of the pre-trained CNN models (without any attention module) we have chosen by-default 10 epochs patience, however, for advance models like Xception, and Convnext we have found the best results come at patience 3. 
    \item Moreover, we have also incorporated adaptive learning rate (alr) [41] in the proposed framework, in which for the first two epochs we maintain fixed lr $1e^{-4}$, thereafter, it will be decaying by a factor of 0.97. 
    \item Furthermore, for the Caltech-256 dataset, we noticed huge over-fitting for any pre-trained CNN model, thus, one set of experiments are also conducted with 50\% of layers kept frozen and only other 50\% layers trained. For training the PiT model, early stopping is avoided. Because of the complexity of the model, the training often stops too early. On the other hand, for SE-Net and CBAM, early stopping criteria with 10 epochs of patience are employed. 
\end{itemize}

\begin{table*}[t]
\begin{center}
\caption{Performance comparisons of several attention frameworks with the proposed framework on caltech-101 and caltech-256 datasets, best two results are presented in bold letters (on testing set)}
\label{tab:my-table}

\begin{tabular}{|c|c|cccc|cccc|}
\hline
\multirow{2}{*}{\begin{tabular}[c]{@{}c@{}}Model/\\ Methods\end{tabular}}          & \multirow{2}{*}{\begin{tabular}[c]{@{}c@{}}Epochs (Ep),\\ ES\end{tabular}} & \multicolumn{4}{c|}{Caltech-101}                                                                                                        & \multicolumn{4}{c|}{Caltech-256}                                                                                                               \\ \cline{3-10} 
                                                                                   &                                                                           & \multicolumn{1}{c|}{Accuracy}       & \multicolumn{1}{c|}{F1score}        & \multicolumn{1}{c|}{AUC}            & Secs/Ep               & \multicolumn{1}{c|}{Accuracy}       & \multicolumn{1}{l|}{F1score}        & \multicolumn{1}{l|}{AUC}            & \multicolumn{1}{l|}{Secs/Ep} \\ \hline
SENet-154 [15]                                                                         & \begin{tabular}[c]{@{}c@{}}10 patience\end{tabular}              & \multicolumn{1}{c|}{0.678}          & \multicolumn{1}{c|}{0.678}          & \multicolumn{1}{c|}{0.965}          &          355             & \multicolumn{1}{c|}{0.461}          & \multicolumn{1}{c|}{0.455}          & \multicolumn{1}{c|}{0.917}          &                1178              \\ \hline
ResNet-152+CBAM                                                                              & \begin{tabular}[c]{@{}c@{}}10 patience\end{tabular}             & \multicolumn{1}{c|}{0.661}          & \multicolumn{1}{c|}{0.653}          & \multicolumn{1}{c|}{0.945}          &         279              & \multicolumn{1}{c|}{0.631}          & \multicolumn{1}{c|}{0.628}          & \multicolumn{1}{c|}{0.963}          &            683                  \\ \hline
\begin{tabular}[c]{@{}c@{}}Novel Pooling\\ by B. Xiao et al. {[}22{]}\end{tabular} & \begin{tabular}[c]{@{}c@{}}50 ep,\\ 10 patience\end{tabular}              & \multicolumn{1}{c|}{0.634}          & \multicolumn{1}{c|}{0.665}          & \multicolumn{1}{c|}{0.944}          &          37             & \multicolumn{1}{c|}{0.377}          & \multicolumn{1}{c|}{0.372}          & \multicolumn{1}{c|}{0.876}          &             83                 \\ \hline
\begin{tabular}[c]{@{}c@{}}Hybrid Attention \\ by J. Pan et al. [23] \end{tabular}       & \begin{tabular}[c]{@{}c@{}}50 ep, \\ 10 patience\end{tabular}             & \multicolumn{1}{c|}{0.608}          & \multicolumn{1}{c|}{0.628}          & \multicolumn{1}{c|}{0.895}          &          175             & \multicolumn{1}{c|}{0.255}          & \multicolumn{1}{c|}{0.271}          & \multicolumn{1}{c|}{0.728}          &              336                \\ \hline
\begin{tabular}[c]{@{}c@{}}Pooling-based\\ Vision Transformer (PiT) \end{tabular}                 & 50 ep                                                                     & \multicolumn{1}{c|}{0.921}          & \multicolumn{1}{c|}{0.919}          & \multicolumn{1}{c|}{0.993}          &          52             & \multicolumn{1}{c|}{0.759}          & \multicolumn{1}{c|}{0.759}          & \multicolumn{1}{c|}{0.974}          &              177                \\ \hline
{\begin{tabular}[c]{@{}c@{}}ConvNext-Tiny \\(pre-trained)\end{tabular}}        & \begin{tabular}[c]{@{}c@{}}50 ep,\\ 3 patience\end{tabular}              & \multicolumn{1}{c|}{0.938}          & \multicolumn{1}{c|}{0.937}          & \multicolumn{1}{c|}{{0.993}} & \multicolumn{1}{c|}{98} & \multicolumn{1}{c|}{{0.825}} & \multicolumn{1}{c|}{{0.826}} & \multicolumn{1}{c|}{{0.978}} &    315                          \\ \hline
\textbf{\begin{tabular}[c]{@{}c@{}}Proposed EAM +\\ DenseNet \end{tabular}}     & \begin{tabular}[c]{@{}c@{}}50 ep,\\ 10 patience\end{tabular}              & \multicolumn{1}{c|}{\textbf{0.952}} & \multicolumn{1}{c|}{\textbf{0.953}} & \multicolumn{1}{c|}{0.993}          & {103}             & \multicolumn{1}{c|}{0.828}          & \multicolumn{1}{c|}{0.838}          & \multicolumn{1}{c|}{\textbf{0.979}}          &             206                 \\ \hline
\textbf{\begin{tabular}[c]{@{}c@{}}Proposed 2EAM +\\ Inception-V3 \end{tabular}}     & \begin{tabular}[c]{@{}c@{}}50 ep,\\ 10 patience\end{tabular}              & \multicolumn{1}{c|}{{0.943}} & \multicolumn{1}{c|}{{0.942}} & \multicolumn{1}{c|}{0.993}          & {142}             & \multicolumn{1}{c|}{\textbf{0.841}}          & \multicolumn{1}{c|}{\textbf{0.846}}          & \multicolumn{1}{c|}{0.967}          &             243                \\ \hline
\textbf{\begin{tabular}[c]{@{}c@{}}Proposed 2EAM +\\ Xception \end{tabular}}     & \begin{tabular}[c]{@{}c@{}}50 ep,\\ 3 patience\end{tabular}              & \multicolumn{1}{c|}{\textbf{0.954}} & \multicolumn{1}{c|}{\textbf{0.955}} & \multicolumn{1}{c|}{0.994}          & {195}             & \multicolumn{1}{c|}{0.835}          & \multicolumn{1}{c|}{0.841}          & \multicolumn{1}{c|}{{0.976}}          &             441                 \\ \hline
\textbf{\begin{tabular}[c]{@{}c@{}}Proposed EAM + \\ ConvNext\end{tabular}}        & \begin{tabular}[c]{@{}c@{}}50 ep,\\ 3 patience\end{tabular}              & \multicolumn{1}{c|}{0.942}          & \multicolumn{1}{c|}{0.944}          & \multicolumn{1}{c|}{\textbf{0.996}} & \multicolumn{1}{c|}{118} & \multicolumn{1}{c|}{\textbf{0.860}} & \multicolumn{1}{c|}{\textbf{0.869}} & \multicolumn{1}{c|}{\textbf{0.989}} &    322                          \\ \hline
\end{tabular}
\end{center}
\end{table*}

\begin{table*}[t]
\begin{center}
\caption{Ablation Studies of the proposed framework for Inception-V3 and DenseNet-121 models, on Caltech datasets. Best result for each model is presented in bold letters (on testing set)}
\label{tab:my-table}

\begin{tabular}{|c|cccc|cccc|}
\hline
\multirow{2}{*}{Models}                                                  & \multicolumn{4}{c|}{Caltech-101 dataset}                                                                                                                                                                                                                           & \multicolumn{4}{c|}{Caltech-256 dataset}                                                                                                                                                                                                   \\ \cline{2-9} 
                                                                         & \multicolumn{1}{c|}{\begin{tabular}[c]{@{}c@{}}Methods with \\ Epochs Patience (EP)\end{tabular}}                            & \multicolumn{1}{c|}{Accuracy}       & \multicolumn{1}{c|}{F1-score}       & \begin{tabular}[c]{@{}c@{}}secs/ \\ epochs\end{tabular} & \multicolumn{1}{c|}{\begin{tabular}[c]{@{}c@{}}Methods with\\ Epochs Patience (EP)\end{tabular}}     & \multicolumn{1}{l|}{Accuracy}       & \multicolumn{1}{c|}{F1-score}       & \begin{tabular}[c]{@{}c@{}}secs/ \\ epochs\end{tabular} \\ \hline
\multirow{4}{*}{\begin{tabular}[c]{@{}c@{}}Inception-\\ V3\end{tabular}} & \multicolumn{1}{c|}{\begin{tabular}[c]{@{}c@{}}10 EP,\\ (100\% fine tuning)\end{tabular}}                                    & \multicolumn{1}{c|}{0.885}          & \multicolumn{1}{c|}{0.887}          & 53                                                      & \multicolumn{1}{c|}{\begin{tabular}[c]{@{}c@{}}10 EP,\\ (100\% fine tuning)\end{tabular}}            & \multicolumn{1}{c|}{0.700}          & \multicolumn{1}{c|}{0.713}          & 216                                                     \\ \cline{2-9} 
                                                                         & \multicolumn{1}{c|}{\begin{tabular}[c]{@{}c@{}}+EAM with 5 EP\\ (100\% fine tinung)\end{tabular}}                            & \multicolumn{1}{c|}{0.914}          & \multicolumn{1}{c|}{0.915}          & 66                                                      & \multicolumn{1}{c|}{\begin{tabular}[c]{@{}c@{}}+EAM 5 EP, \\ (100\% fine tuning)\end{tabular}}       &  \multicolumn{1}{c|}{0.731}          & \multicolumn{1}{c|}{0.740}          & 232                                                     \\ \cline{2-9} 
                                                                         & \multicolumn{1}{c|}{\begin{tabular}[c]{@{}c@{}}+2EAM with 10 EP\\ (100\% fine tuning)\end{tabular}}                          & \multicolumn{1}{c|}{\textbf{0.943}} & \multicolumn{1}{c|}{\textbf{0.942}} & 142                                                        & \multicolumn{1}{c|}{\begin{tabular}[c]{@{}c@{}}+EAM 5 EP, \\ (50\% freeze)\end{tabular}}        & \multicolumn{1}{c|}{0.767}          & \multicolumn{1}{c|}{0.771}          & 170                                                     \\ \cline{2-9} 
                                                                         & \multicolumn{1}{c|}{\begin{tabular}[c]{@{}c@{}}+2EAM with 10 EP\\ (replacing Max-Min\\ pool with Max-pooling)\end{tabular}}  & \multicolumn{1}{c|}{0.924}          & \multicolumn{1}{c|}{0.927}          &   138                                                      & \multicolumn{1}{c|}{\begin{tabular}[c]{@{}c@{}}+2 EAM 10 EP, \\ (50\% freeze)\end{tabular}}     & \multicolumn{1}{c|}{\textbf{0.841}} & \multicolumn{1}{c|}{\textbf{0.846}} & 243                                                        \\ \hline
\multirow{4}{*}{\begin{tabular}[c]{@{}c@{}}DenseNet-\\ 121\end{tabular}} & \multicolumn{1}{c|}{\begin{tabular}[c]{@{}c@{}}10 EP,\\ (100\% fine tuning)\end{tabular}}                                    & \multicolumn{1}{c|}{0.929}          & \multicolumn{1}{c|}{0.932}          & 71                                                      & \multicolumn{1}{c|}{\begin{tabular}[c]{@{}c@{}}10 EP, \\ (100\% fine tuning)\end{tabular}}           & \multicolumn{1}{c|}{0.770}          & \multicolumn{1}{c|}{0.779}          & 429                                                     \\ \cline{2-9} 
                                                                         & \multicolumn{1}{c|}{\begin{tabular}[c]{@{}c@{}}+EAM with 10 EP,\\ (100\% fine tuning)\end{tabular}}                          & \multicolumn{1}{c|}{\textbf{0.952}} & \multicolumn{1}{c|}{\textbf{0.953}} & 103                                                     & \multicolumn{1}{c|}{\begin{tabular}[c]{@{}c@{}}+EAM with 10 EP, \\ (100\% fine tuning)\end{tabular}} & \multicolumn{1}{c|}{0.785}          & \multicolumn{1}{c|}{0.799}          & 317                                                     \\ \cline{2-9} 
                                                                         & \multicolumn{1}{c|}{\begin{tabular}[c]{@{}c@{}}+2EAM with 10 EP,\\ (100\% fine tuning)\end{tabular}}                         & \multicolumn{1}{c|}{0.944}          & \multicolumn{1}{c|}{0.945}          & 130                                                        & \multicolumn{1}{c|}{\begin{tabular}[c]{@{}c@{}}+EAM with 10 EP,\\ (50\% freeze)\end{tabular}}   & \multicolumn{1}{c|}{\textbf{0.828}} & \multicolumn{1}{c|}{\textbf{0.838}} & 206                                                     \\ \cline{2-9} 
                                                                         & \multicolumn{1}{c|}{\begin{tabular}[c]{@{}c@{}}+2EAM with 10 EP\\ (replacing Max-Min \\ pool with Max pooling)\end{tabular}} & \multicolumn{1}{c|}{0.929}          & \multicolumn{1}{c|}{0.930}          &  132                                                       & \multicolumn{1}{c|}{\begin{tabular}[c]{@{}c@{}}+2EAM 10 EP, \\ (50\% freeze)\end{tabular}}      & \multicolumn{1}{c|}{0.819}          & \multicolumn{1}{c|}{0.815}          &  177                                                       \\ \hline
\end{tabular}
\end{center}
\end{table*}

\section{Results and Analysis}
``Results and Analysis'' section can be summarized into main two parts, (a) Performance Comparison of Several existing methods, (b) Validity checking by Explainable AI and 5-fold Cross-Validation. 

\subsection{Performance Comparison}\label{BB}
We have compared the performance of the proposed framework with several existing models in TABLE-I and TABLE-III, across all four datasets. These models include attention-based architectures like SENet-154 [15], ResNet+CBAM [17], Novel Pooling by B.Xiao et al. [22] and Hybrid Attention model by J. Pan et al. [23]. Additionally, we have assessed the performance of two cutting-edge models, (I) ``Pooling-based Vision Transformer (PiT)'', and (II) the latest (pre-trained) CNN model, ``Convext-Tiny'' in TABLE-I. We have demonstrated in TABLE-I that our proposed framework (i.e., EAM$+$DenseNet or, 2EAM$+$Xception) has consistently outperformed all existing models, including the recent trend models, on both Caltech datasets. Only the efficacy of PiT and ConvNext-T are comparable with the performance of the proposed framework, illustrated in TABLE-I. It can be observed from TABLE-I that the proposed framework ``Xception$+$2EAM" achieved the highest accuracy and F1 score of 95.4\% and 95.5\%, respectively, on the Caltech-101 dataset. Moreover, the ConvNext$+$EAM framework has attained a peak accuracy of 86\% and an F1 score of 87\%, on the Caltech-256 dataset. These are the best performances on Caltech datasets so far, to the best of our knowledge. Similarly, from TABLE-III it is apparent that the proposed framework EAM$+$Efficient Net outperformed all existing models as well as it surpassed the efficacy of a recent trend Vision transformer model PiT, by substantial margins, 3\% and 7\% on CIFAR-100 and ImageNet-200 datasets respectively. These results revealed that the proposed framework has worked efficiently in a generalized way across all four datasets. 

\begin{figure}[h]
\centering       \includegraphics[width=8.3cm,height=3.8cm]{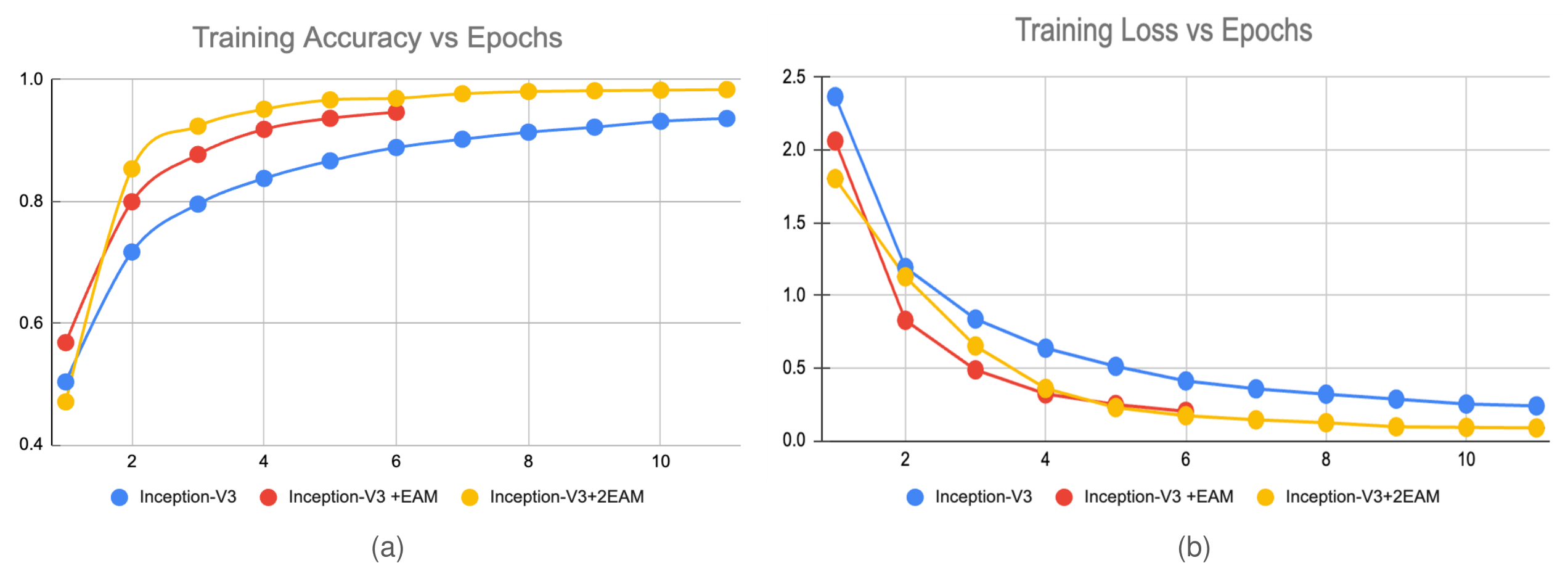}
		\caption{Graph comparison of several frameworks: Inception-V3 (blue color), Inception-V3 +EAM (red-color), Inception-V3 +2EAM (yellow color) on Caltech-256 dataset;  (a). The graph of training Accuracy vs number of epochs, (b). The graph of training loss vs number of epochs}
	\end{figure}

\begin{table*}[t]
\begin{center}
\caption{Performance comparisons of several attention frameworks with the proposed framework on ImageNet-200 and CIFAR-100 datasets, best result is presented in bold letters (on testing set)}
\label{tab:my-table}

\begin{tabular}{|c|c|cccc|cccc|}
\hline
\multirow{2}{*}{\begin{tabular}[c]{@{}c@{}}Model/\\ Methods\end{tabular}}          & \multirow{2}{*}{\begin{tabular}[c]{@{}c@{}}Epochs (Ep),\\ ES\end{tabular}} & \multicolumn{4}{c|}{CIFAR-100 dataset}                                                                                                        & \multicolumn{4}{c|}{ImageNet-200 dataset}                                                                                                               \\ \cline{3-10} 
                                                                                   &                                                                           & \multicolumn{1}{c|}{Accuracy}       & \multicolumn{1}{c|}{F1score}        & \multicolumn{1}{c|}{AUC}            & Secs/Ep               & \multicolumn{1}{c|}{Accuracy}       & \multicolumn{1}{l|}{F1score}        & \multicolumn{1}{l|}{AUC}            & \multicolumn{1}{l|}{Secs/Ep} \\ \hline
SENet-52 [15]                                                                         & \begin{tabular}[c]{@{}c@{}} 5 patience\end{tabular}              & \multicolumn{1}{c|}{0.621}          & \multicolumn{1}{c|}{0.616}          & \multicolumn{1}{c|}{0.971}          &          814             & \multicolumn{1}{c|}{0.263}          & \multicolumn{1}{c|}{0.259}          & \multicolumn{1}{c|}{0.894}          &                827              \\ \hline
ResNet-50+CBAM                                                                              & \begin{tabular}[c]{@{}c@{}}5 patience\end{tabular}             & \multicolumn{1}{c|}{0.585}          & \multicolumn{1}{c|}{0.587}          & \multicolumn{1}{c|}{0.964}          &         619              & \multicolumn{1}{c|}{0.214}          & \multicolumn{1}{c|}{0.207}          & \multicolumn{1}{c|}{0.882}          &            686                  \\ \hline
\begin{tabular}[c]{@{}c@{}}Novel Pooling\\ by B. Xiao et al. {[}22{]}\end{tabular} & \begin{tabular}[c]{@{}c@{}}50 ep,\\ 10 patience\end{tabular}              & \multicolumn{1}{c|}{0.508}          & \multicolumn{1}{c|}{0.498}          & \multicolumn{1}{c|}{0.944}          &          182             & \multicolumn{1}{c|}{0.735}          & \multicolumn{1}{c|}{0.742}          & \multicolumn{1}{c|}{0.944}          &             585                 \\ \hline
\begin{tabular}[c]{@{}c@{}}Hybrid Attention \\ by J. Pan et al. [23] \end{tabular}       & \begin{tabular}[c]{@{}c@{}}50 ep, \\ 10 patience\end{tabular}             & \multicolumn{1}{c|}{0.346}          & \multicolumn{1}{c|}{0.347}          & \multicolumn{1}{c|}{0.895}          &          579            & \multicolumn{1}{c|}{0.101}          & \multicolumn{1}{c|}{0.069}          & \multicolumn{1}{c|}{0.699}          &              315                \\ \hline
\begin{tabular}[c]{@{}c@{}}Pooling-based\\ Vision Transformer (PiT) \end{tabular}                 & 50 ep                                                                     & \multicolumn{1}{c|}{0.781}          & \multicolumn{1}{c|}{0.778}          & \multicolumn{1}{c|}{0.967}          &          179             & \multicolumn{1}{c|}{0.669}          & \multicolumn{1}{c|}{0.668}          & \multicolumn{1}{c|}{0.966}          &              127                \\ \hline
\textbf{\begin{tabular}[c]{@{}c@{}}Proposed EAM +\\ Xception \end{tabular}}     & \begin{tabular}[c]{@{}c@{}}50 ep,\\ 10 patience\end{tabular}              & \multicolumn{1}{c|}{{0.774}} & \multicolumn{1}{c|}{{0.773}} & \multicolumn{1}{c|}{0.994}          & {515}             & \multicolumn{1}{c|}{0.704}          & \multicolumn{1}{c|}{0.707}          & \multicolumn{1}{c|}{0.979}          &             437                \\ \hline
\textbf{\begin{tabular}[c]{@{}c@{}}Proposed EAM +\\ Efficient Net \end{tabular}}     & \begin{tabular}[c]{@{}c@{}}50 ep,\\ 5 patience\end{tabular}              & \multicolumn{1}{c|}{\textbf{0.821}} & \multicolumn{1}{c|}{\textbf{0.825}} & \multicolumn{1}{c|}{0.979}          & {689}             & \multicolumn{1}{c|}{\textbf{0.736}}          & \multicolumn{1}{c|}{\textbf{0.742}}          & \multicolumn{1}{c|}{0.979}          &             585                 \\ \hline
\end{tabular}
\end{center}
\end{table*}
The reason why the proposed framework surpassed the efficacy of all other attention modules is that none of the existing attention models had the ability to provide attention directly to some salient features (i.e., edges) inside the images. To due to this unique capability of extracting edges, the proposed EAM helped the pre-trained CNN models to accelerate its convergence rate, consequently, it also enhanced the model accuracy. This can be further noticed in the graph of Fig.3, where the training accuracy and loss graphs (vs epochs) are presented. This is evident from Fig.3 that after the incorporation of EAM and 2EAMs, Inception-V3 model has converged to less loss much earlier than that of without EAM.
\begin{figure}[h]
		\centering       \includegraphics[width=8.7cm,height=4.4cm]{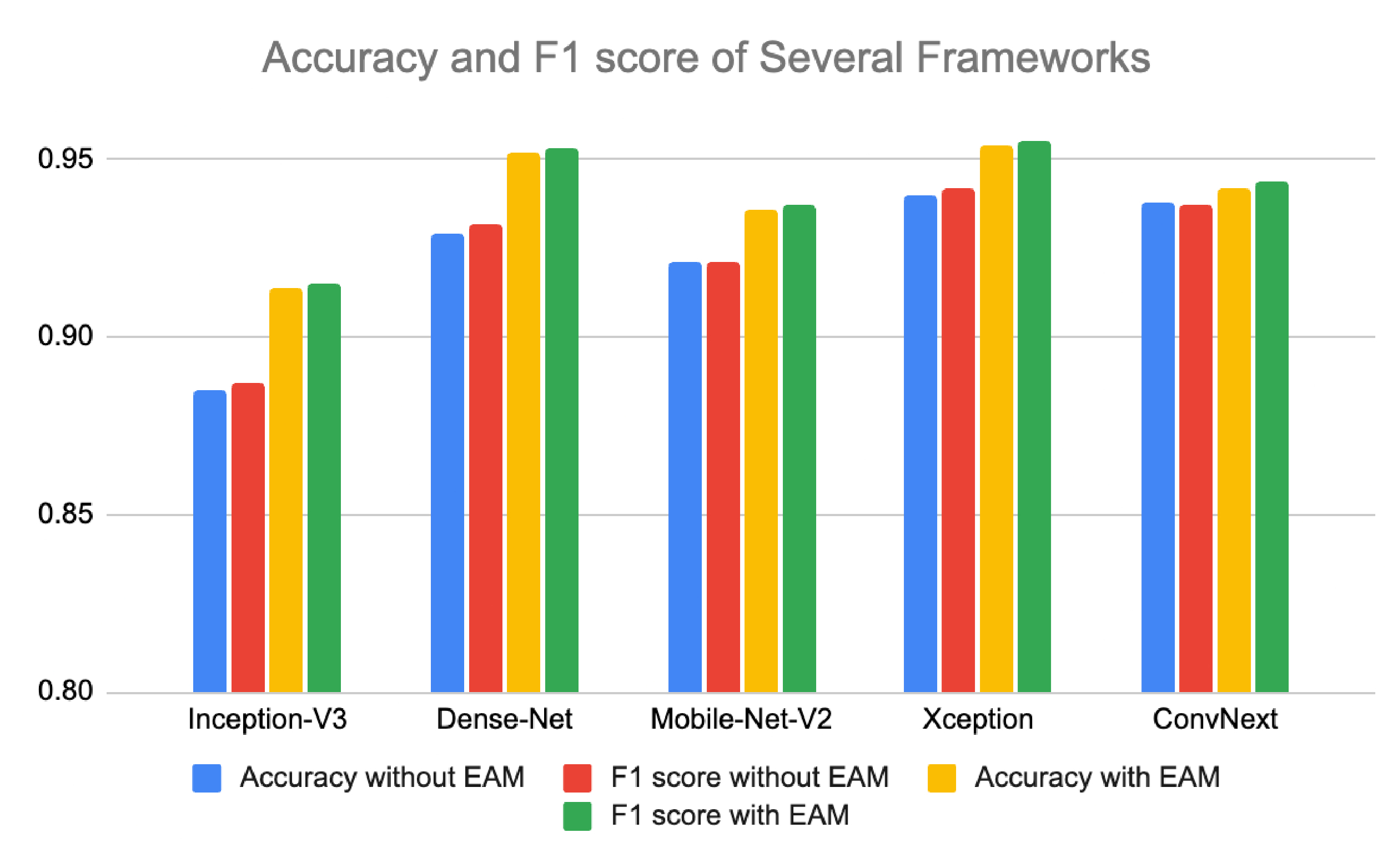}
		\caption{Performance comparison of numerous models with and without EAM on Caltech-101 dataset, all the codes are shared in Github repository for proof}
	\end{figure}

Ablation studies on Caltech-101, Caltech-256 datasets, is presented in TABLE-II. For the experiments of ablation study, we selected only the Inception-V3 [42] and DenseNet-121 [43] models, as the impact of EAM was most noticeable on these models. From this ablation study in TABLE-II, it is quite evident that the proposed attention modules, (that are, EAM and 2EAMs) have boosted the performance of Inception-V3 and DenseNet-121 very consistently. The accuracy and F1 score by the Inception-V3 model have been approximately improved 2.8-3.1\% on both Caltech datasets after the incorporation of single EAM. Furthermore, with the incorporation of 2EAMs on Inception-V3, these accuracy and F1-score are enhanced by 5.8-6.3\% for Caltech datasets. These boosting performances are significant for object classification task, therefore, it justifies the necessity of leveraging the proposed EAM and 2EAM blocks in the pre-trained CNN models. Moreover, this is to clarify that, 2EAM framework does not always outperform the EAM framework as discussed earlier. Therefore, in TABLE-II it can be observed that DenseNet-121 along with 2EAM did not exceed the efficacy of DenseNet-121$+$EAM. However, an important observation is that, both the EAM and 2EAMs framework surpassed the original DenseNet-121 model's performance by 1.5-2.3\% on Caltech-101 and 4-5.8\% on Caltech-256 dataset. This improvement can be visualized prominently in Fig.4. The enhancement of efficacy is more noticeable in Inception-V3 and DenseNet-121 compared to other models such as Xception [44] and ConvNext [40], as depicted in Fig.4. We have observed that due to the use of advanced techniques (and including multiple skip connections), 
\begin{figure}[h]
		\centering	\includegraphics[width=8.7cm,height=3.8cm]{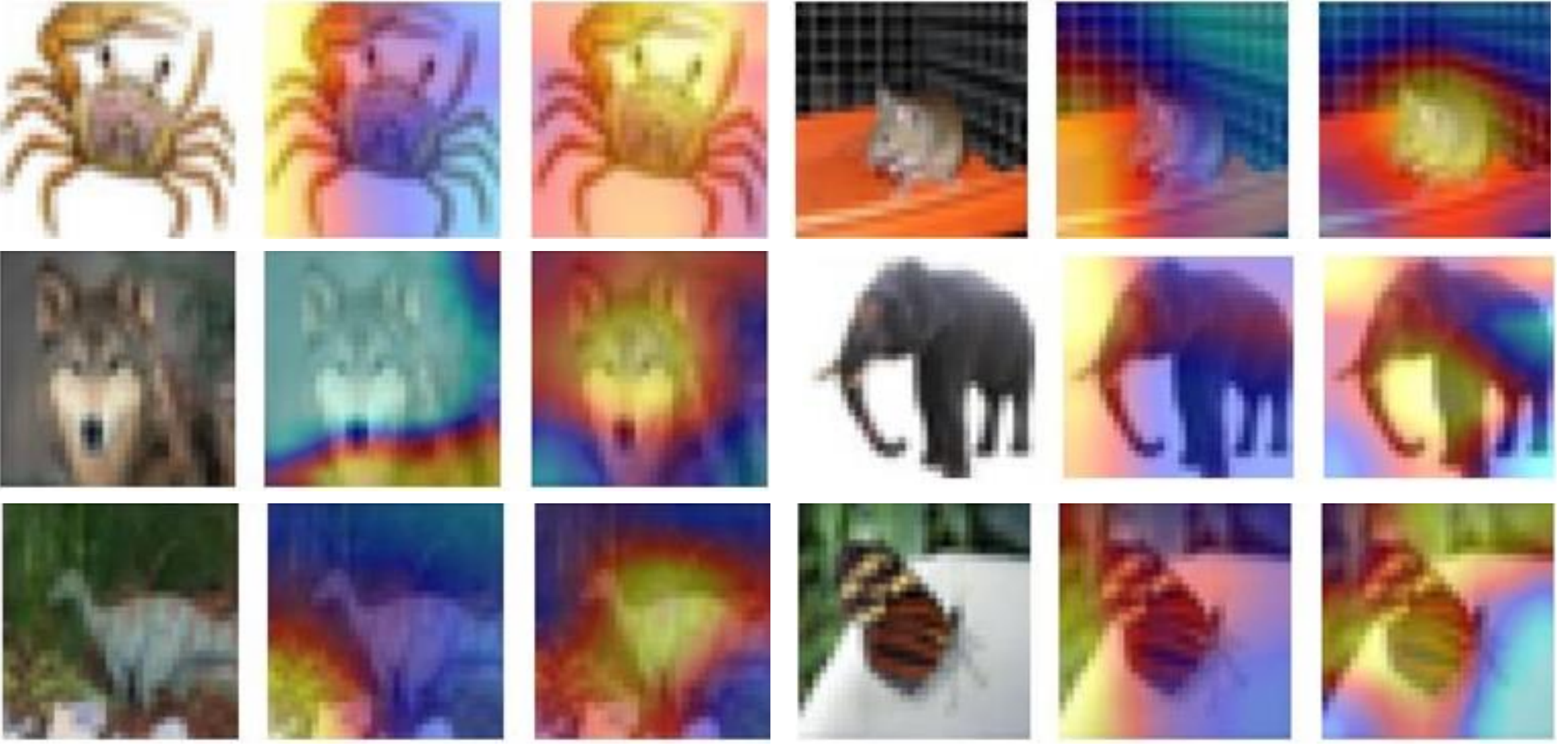}
		\caption{Comparisons of Gradcam heatmaps: $1^{st}$ column of every image represents original images from CIFAR-100 dataset, $2^{nd}$ column represents Gradcam heatmap of Inception-V3 without EAM, $3^{rd}$ column represents Gradcam heatmap of Inception-V3 with 2EAM}
	\end{figure}
the ConvNext and Xception models exhibited smooth convergence during training, and without EAM they have already achieved accuracy of 94\% and 93.8\%, respectively, which are very close to the maximum (saturation level) efficacy on Caltech-101 dataset. Therefore, improvements after incorporating EAM were comparatively minor for the Xception and ConvNext models on Caltech-101.

We conducted a another experiment to evaluate the performance of the 2EAM framework when Max-Min pooling was replaced by Max pooling, given in TABLE-II. This was done to address any potential assumption by the readers that those 2-6\% boosting performance might be attributed to the addition of extra convolutional layers in the EAM module. From Table-II, it is evident that the same 2EAM framework, with Max-Min pooling replaced by Max pooling, did not yield significant improvements for DenseNet and Inception models. Hence, this experiment validates that the main reason behind those 2-6\% improvements (in Table-II) is due to the employment of the Max-Min pooling layer in the EAM block and its ability to effectively capture edge information. Hence, this analysis supports our mathematical analysis given in Section-II-A. 

This is to clarify that due to implementing all these experiments in online platforms (like Google Colab and Kaggle), we could not show any flops in this research. However, from TABLE-I to TABLE-III, we have presented ``secs/ epochs" which indicates how much average time the model needs per epoch during the training. This is analogous to computational complexity of the model. It can be observed from TABLE-I that, other attention frameworks like SENet-52 and CBAM+ResNet-152 had slightly more complexity (higher secs/ ep), compared to our proposed framework. Only the pooling-based ViT (PiT) model had lesser complexity than our framework. However, the complexity of the model is not solely determined by the EAM or 2EAM framework, rather, it primarily depends on the initial architecture of that particular pre-trained CNN model. Moreover, it can be observed that incorporating 2EAMs into any pre-trained CNN model slightly increases its computational complexity. This is due to the incorporation of two Max-Min pooling layers with a 5$\times$5 pool size.

\begin{table}[t]
\begin{center}
\caption{Testing results of the EAM $+$ DenseNet-121, for 5-fold cross validation on Caltech-101 Dataset}
\label{tab:my-table}
\begin{tabular}{|c|c|c|c|c|}
\hline
Folds                                                             & Accuracy                                                        & Precision                                                       & Recall                                                          & F1-score                                                                                                                      \\ 
\hline
fold1                                                             & 0.955                                                           & 0.966                                                           & 0.943                                                           & 0.954                                                                                                                                  \\ \hline
fold2                                                             & 0.946                                                           & 0.967                                                           & 0.935                                                           & 0.950                                                                                                                                  \\ \hline
fold3 \textbf{(max)}                                                           & \textbf{0.959}                                                           & \textbf{0.973}                                                           & \textbf{0.947}                                                           & \textbf{0.959}                                                   \\                                                        
\hline
fold4                                                             & 0.951                                                           & 0.966                                                           & 0.934                                                           & 0.949                                                                                                                                 \\ \hline
fold5                                                             & 0.949                                                           & 0.965                                                           & 0.937                                                           & 0.950                                                                                                                    \\ \hline
\textbf{\begin{tabular}[c]{@{}c@{}}Mean $\pm$\\ Std dev\end{tabular}} & \textbf{\begin{tabular}[c]{@{}c@{}}0.952$\pm$\\ 0.003\end{tabular}} & \textbf{\begin{tabular}[c]{@{}c@{}}0.967$\pm$\\ 0.003\end{tabular}} & \textbf{\begin{tabular}[c]{@{}c@{}}0.939$\pm$\\ 0.004\end{tabular}} & \textbf{\begin{tabular}[c]{@{}c@{}}0.952$\pm$\\ 0.0031\end{tabular}} \\ \hline
\end{tabular}
\end{center}
\end{table}

\subsection{Validity checking of the proposed framework by Explainable AI and cross-validation experiment}
 For checking the validity of the proposed model, we have further performed two more experiments: (I) Stratified 5-fold cross-validation [45], and (II) Explaining the validity by \textit{Explainable AI} [46]. 
\begin{figure}[h]
		\centering	\includegraphics[width=7.4cm,height=6.0cm]{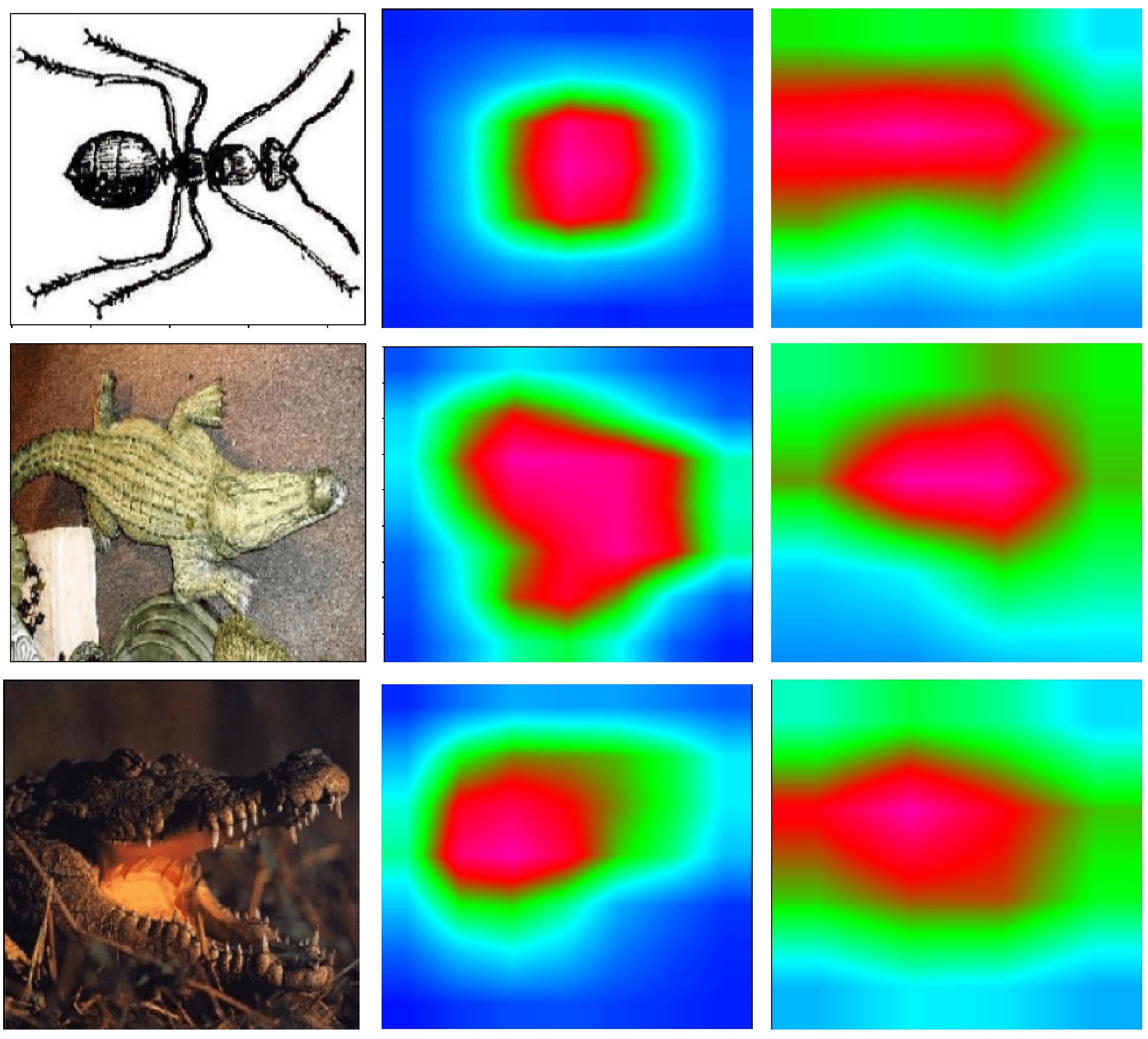}
		\caption{$1^{st}$ column represents original images from Caltech dataset, $2^{nd}$ column represents Gradcam heatmap of DesnseNet without EAM, $3^{rd}$ column represents Gradcam heatmap of DenseNet with EAM}
	\end{figure}

This is to clarify that we have chosen 5-fold cross-validation [46] experiment over 10-fold in Caltech-101 dataset, otherwise in the testing set there would be very few images. As a consequence, there would be large variance between the performance of training and testing sets. We have effectively created the equivalent of 5 different datasets (we call them fold1-to-fold5 in TABLE-IV), where each dataset has distinct testing set, having different statistics compared to the same of other 4 datasets. The splitting of these 5 folds is done in a stratified way, such that the results would not be highly affected by class-imbalance. The results of this 5-fold cross-validation, with mean and standard deviation values, have been presented in Table-IV and also available in the GitHub repository. These results in Table-IV demonstrate that proposed framework (that is DenseNet-121$+$EAM) performed consistently over all the folds. It has achieved an average of accuracy, precision and F1 score 95.2\%, 96.7\% and 95.2\% respectively. Moreover, the standard deviation of these results across all the folds did not exceed 0.4\%, indicating that the proposed framework is highly stable and has generalized well across all 5-folds.

Subsequently, we have  employed Explainable AI, in this subsection. We have presented two distinct Grad-Cam plots [47] in Fig.5 and Fig.6 respectively. In Fig.6, the representation is slightly different than the Fig.5. Here in Fig.6, we have not superimposed the original image on Gradient heatmap such that, overall, its orientation would be very clear. It is evident from Fig.6 that, the heatmap (of Gradients) of the proposed framework is more resilient to the true position and edges of the objects in images compared to the conventional pre-trained frameworks. Moreover, from Fig.5 it is apparent that the heatmap of the proposed framework (2EAMs) predominantly focuses on the objects rather than other regions within the image. Hence, it can be concluded that the proposed EAM or 2EAMs effectively directs the CNN model's attention toward the object edges (or, the areas of interest). This validates the theory of our proposed framework ``EAM'' which was discussed earlier in section II-A and II-B.

\section{Conclusion and Future Work}
A novel ``Edge Attention Module (EAM)'' was proposed for the object classification task. In this EAM branch, one ``Max-Min pooling'' was incorporated for the first time. This pooling was an entirely novel pooling technique which can capture only the edge information effectively. Therefore, this EAM branch provided some additional and essential features about the boundaries (or, strong edges) of the objects that are essential for object classification. Hence, this branch functioned as an edge attention module, enabling the CNN model to focus predominantly on edge information inside the images, and thus, it further enhanced the efficacy of the model. This paper demonstrated that the EAM could be integrated flexibly, simplifying its design part for the users. For instance, in highly challenging scenarios, more number of EAMs could be utilized to further enhance the accuracy substantially. Extensive experimental results also suggested that the proposed EAM, or, 2EAMs along with DenseNet, Xception and ConvNext had outperformed recent trends attention models PiT, CBAM, and SENet-154 by significant margins. Furthermore, a GradCam representation proved that the EAM module enabled the CNN model to focus more on object boundary (or edges), thus, validated our proposed theory on EAM. Previously, this kind of spatial-domain attention module was not available which could directly provide attention to the salient features ``edges". The emergence of the ``Max-Min pooling'' technique has now made this task achievable. Hence, according to our perspective, this was a significant breakthrough which can change the course of employing attention modules in the future, especially in the domain of computer vision. In the future, we are planning to integrate this type of ``edge attention module" or any similar ``problem specific module" into Vision Transformers (ViT) and further want to investigate the effects of these modules on ViT.

\section{Appendix-I}
In this section, we have added an analysis of this Max-Min pooling technique which further strengthens our proposed theory. We have considered two very simple cases to further analyze Max-Min pooling in much deeper. Here in `Case-I' we consider an edgy region (having ramp edge), and in `Case-II' we consider a homogenous region.

\textbf{Case-I}: Let's assume a small portion of an image of size 7$\times$7, associated with edge portion. It has intensity value $I(x,y)$ at location $(x,y)$ given in the following equation (6). This is to clarify that these intensity values in equation (6), are increasing gradually from 109 to 170. This is how the intensity values are likely to vary over the location (x, y) in a digital image, if there is presence of a ramp edge. That means, a sudden (abrupt) change of intensity is not found here. 

\begin{equation}
I(x,y)=\begin{bmatrix}
109 & 110 & 110 & 109 & 110 & 108 & 111\\
114 & 113 & 115 & 114 & 112 & 114 & 113\\
121 & 121 & 121 & 117 & 118 & 121 & 118\\
129 & 128 & 125 & 123 & 121 & 128 & 124\\
141 & 140 & 139 & 136 & 135 & 139 & 131\\
156 & 155 & 151 & 155 & 150 & 153 & 149\\
170 & 169 & 166 & 166 & 163 & 162 & 160\\
\end{bmatrix}
\end{equation}

The most frequent pixel intensity in this matrix is $121$. Now if we apply Max-Min pooling in local window (L, i.e., 5$\times$5) in this matrix, given in equation (6), then we shall get a 2$\times$2 matrix because stride=2 and pool size is 2$\times$2, and according to the formula, 
\begin{equation}
o/p \hspace{0.1cm}size=(\frac{n_h-f_h+2p}{s}+1)\times (\frac{n_v-f_v+2p}{s}+1)
\end{equation}
Here, in equation (7), $n_h$ and $n_v$ are the image size in horizontal and vertical direction respectively; $f_h$ and $f_v$ are the pool size horizontal and vertical direction respectively, `$s$' is the stride. Hence, in this case, the output tensor size after Max-Min pooling will be 
\begin{equation}
o/p \hspace{0.1cm}size=(\frac{7-5+0}{2}+1)\times (\frac{7-5+0}{2}+1)\approx 2\times2
\end{equation}

The first element of this matrix, after employing Max-Min pooling will be $(max value-min value)= 141-109=32$.
\begin{equation}
 {(F_{mn}{(I(x,y))})}_{5\times 5|2}= (141-109)=32 
\end{equation}

Equation (9) can also be analyzed in the following way:
With respect to $121$, there will be a variation in negative direction, i.e., $121-109=12$, and with respect to $121$, there will be a variation in positive direction, i.e., $141-121=20$. Therefore, the total intensity variation is represented by Max-Min pooling operation as follows. 
\begin{equation}
 {(F_{mn}{(I(x,y)_L)})}_{5\times 5|2}= (v_{j,max}+v_{j,min})=(20+12)=32
\end{equation}

This concludes that Max-Min pooling operation has the ability to extract all the intensity deviation present inside a portion of an image, moreover, it exactly matches the mathematical analysis provided in Section II-A. Hence, this analysis justifies our mathematical analysis. 

Eventually, we shall get the output matrix of size 2$\times$2, which is given in the following equation. 

\begin{equation}
F_{mn}({I(x,y)})_{2\times2}=\begin{bmatrix}
32 & 30 \\
53 & 49\\
\end{bmatrix}
\end{equation}
Now applying the conventional Max-Pooling operation of same pool size 5$\times$5 and stride 2 in $I(x,y)$, we might get a completely different matrix, given in equation (12).

\begin{equation}
F_{m}({I(x,y)})_{2\times2}=\begin{bmatrix}
141 & 139 \\
170 & 166\\
\end{bmatrix}
\end{equation}

Comparing equations (11) and (12), we can visualize that the proposed Max-Min pooling does not work like the conventional Max-pooling. From equation (11), it is observed that it could not preserve all intensity information (or, context) inside the image, rather it only senses the intensity variation (or, edge information) inside 5$\times$5 window. If the intensity variation between two pixels inside that window is too high (that is also an indication of strong edge), then Max-Min pooled matrix will generate higher values. On the other hand, Max-pooling in equation (12), tried to preserve the maximum intensity information inside a 5$\times$5 window.

Another observation from equation (12) can be done is that the second row of the matrix in equation (12) shows higher intensity variation than that of first row (which are 53 and 49), which means, there is a possibility that at the bottom side of that image portion, the actual edge is present, therefore, rate of change of intensity is slightly higher in the $2^{nd}$ row than the $1^{st}$ row. 

\vspace{0.1 cm}
\textbf{Case-II:} Now let’s take another matrix (for analysis), $I(x,y)$ in equation (13), where a 7$\times$7 matrix is associated with homogeneous region, means intensity values don't vary too much.

\begin{equation}
I(x,y)=\begin{bmatrix}
109 & 109 & 110 & 109 & 110 & 109 & 110\\
110 & 109 & 111 & 110 & 110 & 110 & 111\\
110 & 110 & 110 & 111 & 111 & 109 & 111\\
111 & 109 & 110 & 109 & 112 & 109 & 110\\
111 & 111 & 111 & 111 & 110 & 112 & 113\\
114 & 112 & 113 & 112 & 113 & 110 & 114\\
116 & 112 & 114 & 116 & 112 & 111 & 117\\
\end{bmatrix}
\end{equation}

After applying Max-Min Pooling with stride 2 and pool size 5$\times$5, we shall get the output matrix of size 2$\times$2, given in the following equation. 
\begin{equation}
F_{mn}({I(x,y)})_{2\times2}=\begin{bmatrix}
3 & 4 \\
7 & 8\\
\end{bmatrix}
\end{equation}
On the other hand, after applying the conventional Max-Pooling operation of same pool size 5$\times$5 and stride 2 in $I(x,y)$ in equation (13), we obtain equation (15).
\begin{equation}
F_{m}({I(x,y)})_{2\times2}=\begin{bmatrix}
112 & 113 \\
116 & 117\\
\end{bmatrix}
\end{equation}

The difference between the conventional Max-pooling and Max-Min pooling is again quite prominent in equations (14) and (15). Furthermore, by comparing equations (11) and (14), it can be concluded that Max-Min pooling may have very low value for homogeneous regions. This value of $3,4,7,8$ could also be reduced to zero, if minimum value exactly equals to maximum value in a small window. Therefore, from this analysis, it is evident that Max-Min pooling may lose some information, if the foreground region is associated with a homogeneous region (just like illustrated in Fig.2, $2^{nd}$ row, in the main manuscript). 

Moreover, it can also be noticed that the last row of equation (14) shows a slight variation in intensity. This minor fluctuation could be attributed to actual intensity changes in a homogeneous region or it could possibly due to noise. Consequently, there is a direct effect on the second row of the (Max-Min pooled) matrix presented in equation (14), where the values in the second row are slightly higher than those in the first row. Hence, it can be concluded that novel \textbf{Max-Min pooling is indeed susceptible to noise. Hence, this analysis further justifies the points 2-3, mentioned in Section II-B}. Furthermore, how to mitigate these limitations of Max-Min pooling, that was already explained in Section-II-B, point 5.

\textbf{Case-III:} Let's extend the case-I, where the $I(x,y)$ is represented by equation (6). However, we deploy a 2$\times$2
Max-Min pooling here instead of 5$\times$5. 

First of all, the size of the Max-Min pooled images now will be changed to 
\begin{equation}
o/p \hspace{0.1cm}size=(\frac{7-2+0}{2}+1)\times (\frac{7-2+0}{2}+1)\approx3\times3
\end{equation}

After applying this Max-Min Pooling with stride 2 and pool size 2$\times$2, we shall get the output matrix of size 3$\times$3, given in the following equation (17).   

\begin{equation}
F_{mn}({I(x,y)})_{3\times3}=\begin{bmatrix}
5 & 6 & 6 \\
8 & 8 &  10\\
16 & 19 & 18 \\
\end{bmatrix}
\end{equation}

Comparing equations (11) and (17), we can observe that if Max-Min pooling is applied with a relatively small pool size of 2$\times$2, it tends to extract very low values in edgy regions compared to the same with pool size 
5$\times$5. This reveals that strong edges cannot be effectively captured while utilizing a small pool size for Max-Min pooling. Therefore, in this research, we have chosen a large pool size of 5$\times$5 in Max-Min pooling, to ensure that there will be extraction of strong edges. Because it is likely that inside a 5$\times$5 window, there will be more intensity variation than that of 2$\times$2 window. It can be further verified from the Fig.2, column 3 and column 4. From this Fig.2, it is evident that Max-Min pooled image (in column 3, with 2$\times$2 pool size) lost some significant features and most of the intensity values are very low, hence, in this case we obtain dark images. In the contrary, when the pool size in Max-Min pooling is increased to 5$\times$5 (Fig.2, column 4), then it captures strong edges efficiently within the image. This analysis further explains why Max-Min pooling with pool size 5$\times$5 is preferable over the pool size of 2$\times$2. \textbf{Hence, it justifies the $1^{st}$ property of EAM, mentioned in Section II-B.}

\end{document}